\documentclass[11pt,twocolumn,twoside]{IEEEtran}
\usepackage{amsmath}
\usepackage[margin=0.75in,headheight=0.45in]{geometry}
\usepackage[pdftex]{epsfig}
\usepackage{amsfonts}
\usepackage{amssymb}
\usepackage{fancyhdr}
\usepackage{url}
\usepackage{amsmath}
\usepackage{cite}

\include{graphicsx}

\begin{document}
	\title{\vspace{0.2in}\sc GlobeNet: Convolutional Neural Networks for Typhoon Eye Tracking from Remote Sensing Imagery}
	\author{Seungkyun Hong$^{*,1,2}$\thanks{$^1$Korea University of Science and Technology (UST) $^2$Korea Institute of Science and Technology Information (KISTI) $^\dagger$Corresponding Author}, Seongchan Kim$^{2}$, Minsu Joh$^{1,2}$, Sa-kwang Song$^{\dagger,1,2}$}
    
	\maketitle
	\thispagestyle{fancy}
	\begin{abstract}
		Advances in remote sensing technologies have made it possible to use high-resolution visual data for weather observation and forecasting tasks. We propose the use of multi-layer neural networks for understanding complex atmospheric dynamics based on multi-channel satellite images. The capability of our model was evaluated by using a linear regression task for single typhoon coordinates prediction. A specific combination of models and different activation policies enabled us to obtain an interesting prediction result in the northeastern hemisphere (ENH).
	\end{abstract}
	
	\section{Introduction}
	Recent decades have seen significant efforts by meteorologists to develop numerical weather prediction (NWP) models such as Weather Research\&Forecasting (WRF) to predict and produce rich atmospheric metrics such as the air pressure, temperature, and wind speed. The purpose of these processes is to predict extreme weather events capable of causing severe damage to human society. Typhoons, i.e., mature tropical cyclones known to commonly develop in the northwestern Pacific Basin, are one of the targets of atmospheric dynamics modeling. However, these models require considerable computational resources and processing time.
	
	Previous studies \cite{kordmahallehhurricane,racah2016semi,xingjian2015convolutional} have shown that deep neural networks yield reliable results for weather-related problems and are computationally less intensive compared to large NWP models. Recent weather research comprising 3D data, such as weather simulation results or radar reflectivity datasets, involved the application of convolutional neural networks (CNN) which are known to be capable of extracting rich regional features from multi-dimensional data.
	
	Meanwhile, advances in satellite equipment have made it possible to accumulate extensive global observations than was previously the case. An imager, which is one of the core instruments that are used as satellite components, collects visual global observations coupled with infrared (IR) and visible (VIS) wavelength sensing. Modern satellite imagers have capabilities of 0.25\textasciitilde4km at s.s.p. (Spatial resolution) and multiple channels from 5ch (MI) to 36ch (MODIS). Considering that high-resolution global observations exceeding 1 TB in size are collected by several weather research centers daily, it has become possible to use sophisticated visual information from massive datasets.
	
	Nevertheless, an approach for typhoon trajectory tracking based on bare remote sensing images has not been reported yet. Moreover, high-resolution imagery itself has never been utilized extensively without any modification on information. Our solution to these problems was to focus on the utilization of the entire visual context from large-scale global observation to yield models for typhoon eye tracking based on deep CNNs. We first present two discrete neural networks based on multiple convolutional layers to develop an understanding of complex atmospheric dynamics, and then discuss the prediction results.
	
	\section{Related work}
	In recent years, many researchers have investigated the use of deep neural networks to solve various weather-related problems. Kordmahalleh et al. \cite{kordmahallehhurricane} explored a model to predict cyclone tracks from the large NOAA best track database. Racah et al. \cite{racah2016semi} suggested a semi-supervised model for extreme weather events from long-term CAM5 climate simulation datasets. Xingjian et al. \cite{xingjian2015convolutional} suggested a complex CNN-LSTM network (ConvLSTM) for the prediction of future precipitation rates from reflectivity data recorded by ground-based radar stations. However, not many investigations involved processing high-dimensional imagery data of weather phenomena.
	
	Several complex network topologies proved capable of high-accuracy prediction or classification such as image classification from multi-layer inception followed by repeated convolution-pooling steps \cite{szegedy2015going} and symmetric skip connections for clear image restorations \cite{mao2016image}. Unlike other CNN surveys, our research focuses on the characteristics of high-resolution remote sensing images containing vast and detailed descriptions of typhoon and cloud movement. In other words, our models utilize satellite images by preserving their details without significant modification due to processing such as image resizing or cropping.
	
	\section{Method}
	\subsection{Data}
	
	\begin{figure}
		\begin{center}
			\epsfxsize=0.95\hsize \epsfbox{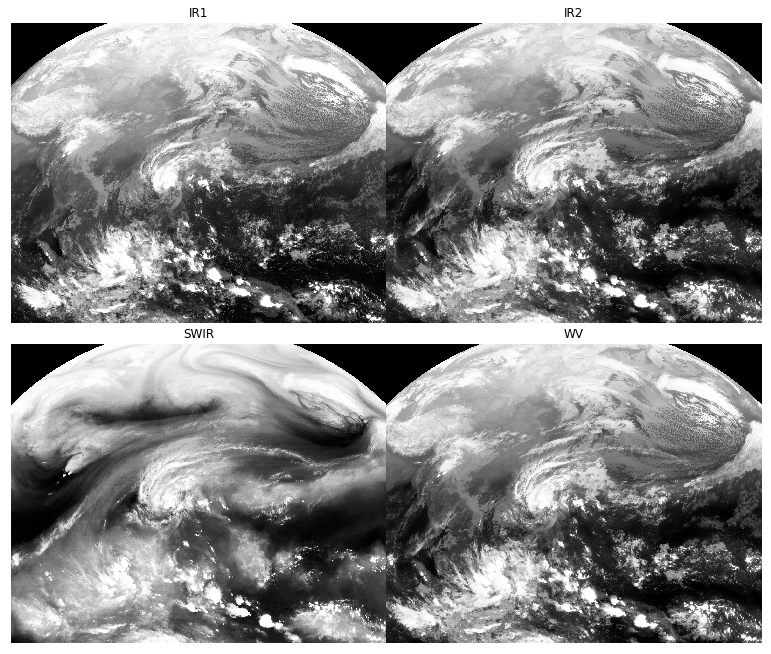}
		\end{center}
        \vspace*{-0.5cm}
		\caption{COMS-1 MI 4ch IR images with CDF Histogram Normalization}
		\label{fig:coms_mi_samples}
        \vspace*{-0.25cm}
	\end{figure}
	
\begin{figure}[h]
		\begin{center}
			\epsfxsize=0.95\hsize \epsfbox{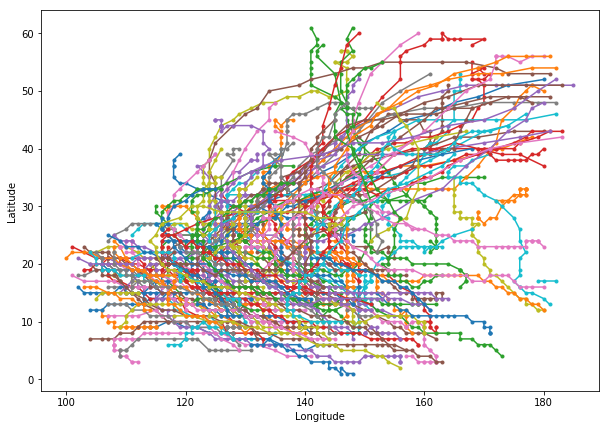}
		\end{center}
        \vspace*{-0.5cm}
		\caption{Typhoon trajectories spanning 6 years (2011\textasciitilde2016) from the JMA RMSC-Tokyo Best Track Data} 
		\label{fig:best_track}
        \vspace*{-0.25cm}
	\end{figure}

	Typhoon track prediction from remote sensing images requires two detailed types of information: a trajectory point consisting of the latitude and longitude and a single satellite image. The trajectory dataset we used was the Japan Meteorological Agency (JMA)'s official best track information \textbf{(Figure \ref{fig:best_track})}, which was used with a decimal precision of 1. The satellite image set comprised images acquired with the COMS-1 \cite{ou2005introduction} meteorological imager (MI) of the Korea Meteorological Agency (KMA). The MI covers five channels including four IR images and one VIS image; however, because VIS imaging cannot be used for observation of the area of interest at midnight, only the 4-ch IR images \textbf{(Figure \ref{fig:coms_mi_samples})} were chosen for the image dataset.
	
	We used 2,674 satellite images collected from April 1, 2011 until the end of 2016, covering nearly six years and 152 occurrences of single typhoons. The scope of the scale level between the latitude (80$^{\circ}$) and longitude (150$^{\circ}$) was matched by normalizing both the input (satellite image) and output (typhoon track with latitude and longitude) to a value between 0 and 1.
	Furthermore, both the image and track data are picked with a small size and reshaped immediately before model input for mini-batch sampling. Therefore, the dimensionality of an image becomes four, denoted by the $NumSamples \times Height \times Width \times Channels$ (NHWC) format, and that of the track becomes three ($NumSample \times Latitude \times Longitude$). 
	
	\subsection{Models}
	The model receives 3D satellite images denoted by NHWC format as inputs. Our research surveyed two discrete network topologies for extracting rich features of cloud shapes. After an input image passes the multiple convolutional layers, each network has the exact fully connected dense layers for linear metrics regression. In the regression step, any values related to the weather event can be the target value. Our networks are developed to achieve fast examination; thus, they only fit a point of the typhoon from the input image.
	The overall prediction process can be described as follows:
	\begin{figure*}[t]
		\center
		\includegraphics[width=\textwidth]{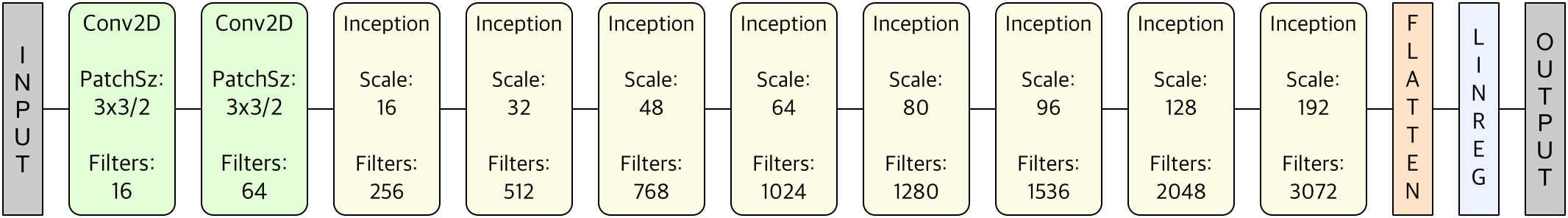}
		\caption{Network Topology of Complex CNN with Inception Units}
		\label{fig:regressor_cplx}
        \vspace*{-0.5cm}
	\end{figure*}
	\begin{enumerate}
		\item An input image is used for feature extraction by multiple convolutional filters, followed by application of the max-pooling technique. Each model has a different convolution policy (basic vs. inception unit)
		\item Fully connected layers following flattening of the filtered images builds a nonlinear connection for predicting the point of a single typhoon eye.
		\item A smaller dense layer is used to ensure that the model fits its final outcome and ground truth, stimulated by linear activation.
	\end{enumerate}
   	\textit{More detailed model descriptions are below:}
	\subsubsection{Simple CNN with Fast Striding}
\begin{figure}[h]
    	\vspace*{-0.5cm}
		\begin{center}
			\epsfxsize=0.95\hsize \epsfbox{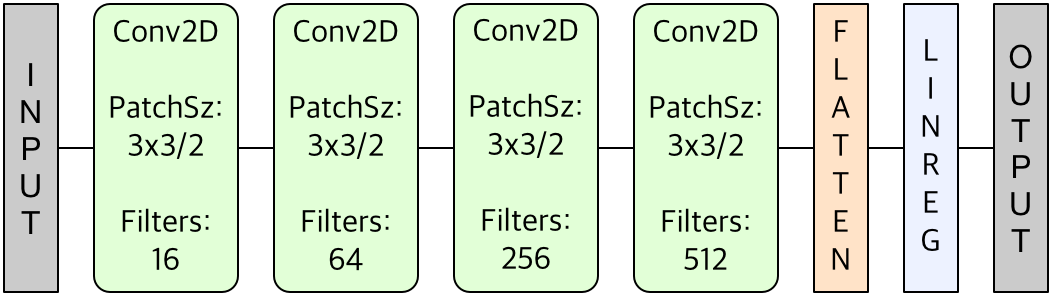}
		\end{center}
        \vspace*{-0.25cm}
		\caption{Network Topology of Simple CNN with Fast Striding}
		\label{fig:regressor}
        \vspace*{-0.25cm}
	\end{figure}

	Our first model \textbf{(Figure \ref{fig:regressor})} uses four conv. layers in conjunction with max pooling, which is similar to the basic CNN named LeNet-5 \cite{lecun1995learning}. Because of the high dimensionality of the input data, we increased the strides from 1 to 2 when applying the filters. The larger strides resulted in each layer extracting fewer sparse features, further minimizing computational resources. After each image passes the conv. layers, max pooling is performed. This enhances the value of features and reduces the data size.
    \subsubsection{Complex CNN with Joint Inception Units}
	Our second model \textbf{(Figure \ref{fig:regressor_cplx})} starts by jointly using two conv. layers \& max pooling steps for image reduction and as candidates to prepare for feature extraction. Then, eight deep-and-complex inception layers follow previously convolved images similar to the approach used in GoogLeNet \cite{szegedy2015going}. In each inception unit, the images pass four different filter policies involving various types of feature extraction. After each convolutional procedure in the inception unit has been completed, all the filtered images are concatenated to produce the same image dimension with increased depth.
	\begin{figure}[h]
		\begin{center}
			\epsfxsize=0.95\hsize \epsfbox{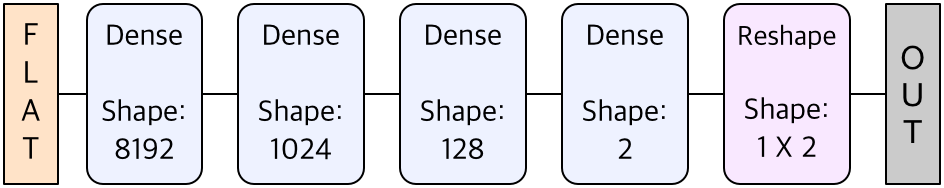}
		\end{center}
        \vspace*{-0.25cm}
		\caption{Fully connected layers for linear regression process after flattening the convolutional layer}
		\label{fig:linreg}
        \vspace*{-0.1cm}
	\end{figure}
	\subsubsection{Linear metrics regression}
	After the CNN has extracted many image features from an original input image, the network produces low-dimensional images with greater channel depth. These deep images are flattened to build fully connected (FC) dense layers. \textbf{(Figure \ref{fig:linreg})} Three fixed FC layers develop a nonlinear connection for obscure value prediction. Finally, a single FC layer with a sigmoid activation function produces the predicted outcome coupled with a linear activation unit for linear regression.
	
	\section {Evaluation}
	The accuracy of typhoon eye tracking models is defined as (\ref{exp:rmse}), where \textit{P} (\ref{exp:desc_point_gt}) is a point consisting of the latitude and longitude of the ground truth and \textit{$\hat{P}$} (\ref{exp:desc_point_pred}) is a point from the prediction result:
	\begin{equation}
	RMSE_{Prediction} = \sqrt[]{\frac{1}{N} \sum_{n=1}^{N} {(P-\hat{P})}^2}
	\label{exp:rmse}
	\end{equation}
	\begin{equation}
	P = (Lat_{gt}, Long_{gt})
	\label{exp:desc_point_gt}
	\end{equation}
	\begin{equation}
	\hat{P} = (Lat_{pred}, Long_{pred})
	\label{exp:desc_point_pred}
	\end{equation}
	
	We examined the accuracy of our models for linear regression by conducting multiple experiments with several different configurations. Each network can set two different activation functions for each convolutional step (ReLU/LeakyReLU \cite{maas2013rectifier}/ELU \cite{clevert2015fast}) and fully connected step (Sigmoid/Tanh). The Adam optimizer \cite{kingma2014adam} is used for gradient optimization with an initial learning rate 1e-5, which is known to achieve fast optimization. The entire dataset is divided into training and testing sets in the ratio 9:1.
    
	All our models use the toolkit named Keras \cite{chollet2015keras} with the TensorFlow \cite{abadi2016tensorflow} backend as a neural network framework.
	
	\begin{center}
		\textbf{Testbed Environment Configuration}
		\begin{itemize}
			\item CPU: Intel\textsuperscript{\tiny\textregistered} Xeon\textsuperscript{\tiny\textregistered} E5-2660v3 @ 2.60 GHz
			\item GPU: NVIDIA\textsuperscript{\tiny\textregistered} Tesla\textsuperscript{\tiny\texttrademark} K40m 12 GB
			\item Platform: Python 3.5.2
			\item Core: Keras 2.0.5 with TensorFlow 1.2
		\end{itemize}
	\end{center}
	\begin{figure}[h]
		\begin{center}
			\epsfxsize=0.95\hsize \epsfbox{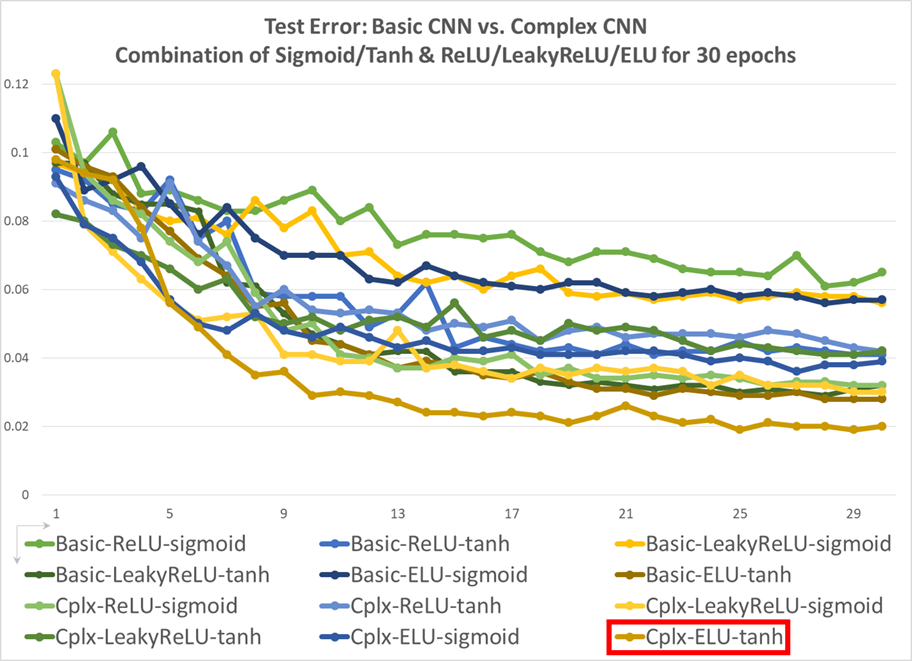}
		\end{center}
		\caption{Test Error: Basic CNN vs. Complex CNN - Combination of ReLU/LeakyReLU/ELU \& Sigmoid/Tanh}
		\label{fig:test_error}
        \vspace*{-0.5cm}
	\end{figure}
	
	\textbf{Figure \ref{fig:test_error}} shows the overall test error on the ``test data’’ after 30 epochs of training with the training data. The best prediction was achieved with Complex CNN with ELU/Tanh activation after Conv2D/Dense layers, with an RMSE of 0.02, about 74.53 km in great circle distance. In contrast, the worst prediction models was Basic CNN with ReLU/sigmoid activation, characterized by an RMSE of 0.065 which is about 362.91 km in great circle distance.  Every result obtained with Complex CNN was more accurate than with Simple CNN with the same activation combinations.
    
    However, only half the number of samples was used to train Complex CNN compared to Basic CNN because of the massive amount of memory that was used. The large memory requirement slows down training and requires more processing time for Complex CNN relative to Basic CNN. Nevertheless, once the models were built, feed-forward always finished within a few seconds on both networks.
	
	\section{Conclusion and Future works}
	We proposed two different neural networks to develop an understanding of atmospheric dynamics and weather events, for typhoons in particular. We studied typhoon tracking by attaching the observation images after evaluating the capability of the model to extract the topology features. This produced an RMSE of approximately 0.02 to guess the center of a single typhoon in a space of 80-degrees latitude by 150-degrees longitude.
	
	Moreover, a particular combination of activation between convolutional filters and dense layers provided the best results in the experiments (exponential linear unit \& hyperbolic tangent). We further investigated representations and prediction of visual weather events by the pre-trained network.
	
	However, our model only focused on single-event typhoons in remote sensing images. Hence, we also plan to survey multiple occurrences in a single image that would require simultaneous tracking, as well as predicting additional weather metrics such as the precipitation, air pressure, temperature, and wind speed for the simulation of complex atmospheric circulation and long-term global climate modeling.
	
	\section*{Acknowledgments}
	This work formed part of research projects carried out at the Korea Institute of Science and Technology Information (KISTI) (Project No. K-17-L05-C08, Research for Typhoon Track Prediction using an End-to-End Deep Learning Technique)
	
	\bibliographystyle{ieeetr}
	\bibliography{ci2017.bib}

\begin{thebibliography}{10}

\bibitem{kordmahallehhurricane}
M.~M. Kordmahalleh, M.~G. Sefidmazgi, A.~Homaifar, and S.~Liess, ``Hurricane
  trajectory prediction via a sparse recurrent neural network,''

\bibitem{racah2016semi}
E.~Racah, C.~Beckham, T.~Maharaj, C.~Pal, {\em et~al.}, ``Semi-supervised
  detection of extreme weather events in large climate datasets,'' {\em arXiv
  preprint arXiv:1612.02095}, 2016.

\bibitem{xingjian2015convolutional}
S.~Xingjian, Z.~Chen, H.~Wang, D.-Y. Yeung, W.-K. Wong, and W.-c. Woo,
  ``Convolutional lstm network: A machine learning approach for precipitation
  nowcasting,'' in {\em Advances in neural information processing systems},
  pp.~802--810, 2015.

\bibitem{szegedy2015going}
C.~Szegedy, W.~Liu, Y.~Jia, P.~Sermanet, S.~Reed, D.~Anguelov, D.~Erhan,
  V.~Vanhoucke, and A.~Rabinovich, ``Going deeper with convolutions,'' in {\em
  Proceedings of the IEEE conference on computer vision and pattern
  recognition}, pp.~1--9, 2015.

\bibitem{mao2016image}
X.~Mao, C.~Shen, and Y.-B. Yang, ``Image restoration using very deep
  convolutional encoder-decoder networks with symmetric skip connections,'' in
  {\em Advances in Neural Information Processing Systems}, pp.~2802--2810,
  2016.

\bibitem{ou2005introduction}
M.-L. Ou, S.-R.~C. Jae-Gwang-Won, {\em et~al.}, ``Introduction to the coms
  program and its application to meteorological services of korea,'' in {\em
  Proceedings of the 2005 EUMETSAT Meteorological Satellite Conference,
  Dubrovnik, Croatia}, pp.~19--23, 2005.

\bibitem{lecun1995learning}
Y.~LeCun, L.~Jackel, L.~Bottou, C.~Cortes, J.~S. Denker, H.~Drucker, I.~Guyon,
  U.~Muller, E.~Sackinger, P.~Simard, {\em et~al.}, ``Learning algorithms for
  classification: A comparison on handwritten digit recognition,'' {\em Neural
  networks: the statistical mechanics perspective}, vol.~261, p.~276, 1995.

\bibitem{maas2013rectifier}
A.~L. Maas, A.~Y. Hannun, and A.~Y. Ng, ``Rectifier nonlinearities improve
  neural network acoustic models,'' in {\em Proc. ICML}, vol.~30, 2013.

\bibitem{clevert2015fast}
D.-A. Clevert, T.~Unterthiner, and S.~Hochreiter, ``Fast and accurate deep
  network learning by exponential linear units (elus),'' {\em arXiv preprint
  arXiv:1511.07289}, 2015.

\bibitem{kingma2014adam}
D.~Kingma and J.~Ba, ``Adam: A method for stochastic optimization,'' {\em arXiv
  preprint arXiv:1412.6980}, 2014.

\bibitem{chollet2015keras}
F.~Chollet {\em et~al.}, ``Keras.'' \url{https://github.com/fchollet/keras},
  2015.

\bibitem{abadi2016tensorflow}
M.~Abadi, P.~Barham, J.~Chen, Z.~Chen, A.~Davis, J.~Dean, M.~Devin,
  S.~Ghemawat, G.~Irving, M.~Isard, {\em et~al.}, ``Tensorflow: A system for
  large-scale machine learning.,'' in {\em OSDI}, vol.~16, pp.~265--283, 2016.

\end{thebibliography}
	
\end{document}